\documentclass[10pt,twocolumn,letterpaper]{article}

\usepackage{cvpr}              

\usepackage{enumitem}
\usepackage{subcaption}
\usepackage{multicol}
\usepackage{multirow}
\usepackage{colortbl}
\usepackage{amsmath}
\usepackage{amssymb}
\usepackage{csquotes}
\definecolor{LightCyan}{rgb}{0.88,1,1}
\definecolor{Pearl}{HTML}{F1EAE3}
\usepackage{booktabs}

\definecolor{cvprblue}{rgb}{0.21,0.49,0.74}
\usepackage[pagebackref,breaklinks,colorlinks,allcolors=cvprblue]{hyperref}

\def\paperID{17685} 
\def\confName{CVPR}
\def\confYear{2025}

\title{SC3D: Label-Efficient Outdoor 3D Object Detection via Single Click Annotation}

\author{Qiming Xia$^1$ \quad Hongwei lin$^{1}$ \quad Wei Ye$^{1}$ \quad Hai Wu$^{1}$ \quad Yadan Luo$^2$ \quad Cheng Wang$^1$ \quad Chenglu Wen$^1$\thanks{Corresponding author}\\
$^1$Xiamen University, Xiamen, China\\ \quad $^2$University of Queensland, Australia\\
}

\begin{document}
\maketitle
\begin{abstract}
LiDAR-based outdoor 3D object detection has received widespread attention. However, training 3D detectors from the LiDAR point cloud typically relies on expensive bounding box annotations. This paper presents \textbf{SC3D}, an innovative label-efficient method requiring only a single coarse click on the bird's eye view of the 3D point cloud for each frame.
A key challenge here is the absence of complete geometric descriptions of the target objects from such simple click annotations.
To address this issue, our proposed \textbf{SC3D} adopts a progressive pipeline.
Initially, we design a mixed pseudo-label generation module that expands limited click annotations into a mixture of bounding box and semantic mask supervision.
Next, we propose a mix-supervised teacher model, enabling the detector to learn mixed supervision information. Finally, we introduce a mixed-supervised student network that leverages the teacher model's generalization ability to learn unclicked instances.
Experimental results on the widely used nuScenes and KITTI datasets demonstrate that our \textbf{SC3D} with only coarse clicks, which requires only  0.2\% annotation cost, achieves state-of-the-art performance compared to weakly-supervised 3D detection methods.
The code will be made publicly available.
\end{abstract}    
\section{Introduction}
\label{sec:intro}
\begin{figure}[t]
  \centering
   \includegraphics[width=0.99\linewidth]{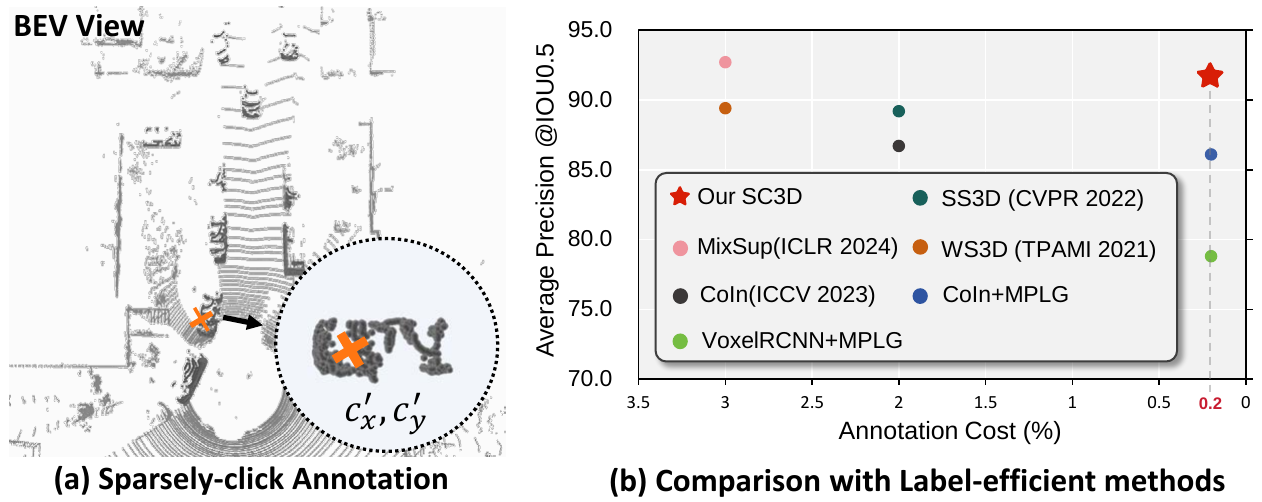}
   \caption{(a) Unlike traditional costly box annotations, coarse-click alternatives require only quick clicks on single object in the 2D BEV plane, yet offer limited supervision; (b) demonstrates a comparison with existing label-efficient methods on the KITTI dataset, where we reduce the annotation cost to 0.2\%, enhancing the previous scheme's labeling efficiency by over 10 times, while still maintaining the performance of the detector. }
   \label{fig:intro}
\end{figure}
In recent years, notable progress has been made in LiDAR-based 3D object detection research \citep{virtual, rain_det, VoxelNext, DSVT}. Despite these advancements, the need for precise bounding box supervision remains a major challenge due to its time-consuming and labor-intensive nature. For instance, the KITTI dataset~\citep{KITTI} contains 3,712 training scenes with over 15,000 vehicle instances, where manual annotation of a single instance can take roughly 114 seconds~\citep{WS3D}. 
The extensive labeling efforts required escalate dramatically when scaling detectors to larger-scale datasets~\cite{waymo, nuScenes, ONCE}, thereby hindering further research in a fully supervised manner.

To alleviate the annotation burden, recent studies have explored alternatives that require \textit{fewer} annotated frames or instances to train high-performing 3D object detectors. Specifically, \textit{semi-supervised methods}~\citep{3DIoUMatch, detmatch, hssda} leverage a subset of the annotated frames, while 
\textit{sparsely-supervised} methods \citep{CoIn, ss3d, HINTED} rely on only one bounding box annotation per frame during training. While these approaches have significantly lowered annotation costs, annotating 6DoF bounding boxes for every scene remains time-consuming. 

To provide a faster, albeit less precise, method of human supervision, 
WS3D~\citep{WS3D} and ViT-WSS3D \citep{WSS3D} propose a \textit{mixed supervision} strategy that replaces some box annotations with the \textbf{center-click} annotations. In this approach, annotators click the center of objects on the Bird's Eye View (BEV) to generate center-level labels, reducing labeling time per instance to approximately 2.5 seconds, which is 50 times faster than traditional bounding box labeling. 

However, relying on full coverage of the central point has significant limitations: (1) it requires annotators to precisely indicate the \textit{center} position for all instance in each frame and (2) it fails to accurately represent the \textit{shape} and \textit{scale} of objects, especially in sparse point clouds where objects may be partially observed. These challenges become even more severe for \textit{moving objects}, where motion across frames further complicates the estimation of an accurate bounding box from clicks. As a result, prior works have struggled to scale up the use of sparse click supervision, combined with traditional box annotation for mixed supervision, which diminishes the overall effectiveness. 

In this paper, we introduce a label-efficient 3D object detection approach, SC3D, which performs only sparse click annotations (Fig.~\ref{fig:intro}), greatly reducing the annotation cost.
Our framework comprises three key designs: (1) To recover accurate supervisions from coarse click annotations, we use temporal cues to determine the status of the clicked instances and then generate mixed-supervisions. (2) To optimize learning from mixed supervision, we designed a mixed-supervised teacher network to capture the fundamental patterns within the mixed supervision data. (3) To learn from unclicked instances, we introduce a mixed-supervised student network that uses the teacher's generalization to capture their patterns.
We evaluated SC3D on the widely adopted nuScenes and KITTI datasets. Remarkably, SC3D achieves competitive performance with weakly-supervised baselines that rely on accurate box annotations, demonstrating the effectiveness of our approach under sparely click conditions. In summary, our contributions are:
\begin{itemize}[leftmargin=10pt]

\item We propose the first method of sparse click annotated outdoor 3D object detection (SC3D), which solely relies on a single coarse click on each frame. This approach dramatically reduces the annotation cost of 3D object detection tasks to 0.2\%. 

\item We design a mixed pseudo-label generation module that combines temporal cues and local point cloud distribution to recover accurate supervision from coarse click annotations.

\item We propose a mixed-supervised teacher-student network that expands the information of unclicked instances from limited mixed supervision, thereby enhancing the performance of the detector.

\end{itemize}

\section{Related Work}

\paragraph{LIDAR-based 3D Object Detection.} In recent years, fully-supervised 3D object detection has been widely studied. The early methods \citep{PointPillars, SECOND, CenterPoint, Voxelnet, SA-SSD} utilized an end-to-end one-stage object detection strategy, predicting detection boxes directly from point clouds. The advantage of these approaches is that the detector inference speed is fast, which more easily meets the real-time requirements of the autonomous driving. To improve the performance of 3D detector, the two-stage methods \citep{PointRCNN, Voxel-RCNN, PV-RCNN++, TED, VoxelNext, club, bevfusion, DSVT}
introduced an additional proposal refinement stage, which improves detection performance by refining regions of interest. 
Meanwhile, multi-stage methods \citep{CasA, 3dcas} progressively refine detection proposals through a series of stages, each employing a cascaded optimization strategy. This iterative process leads to increasingly accurate predictions as the algorithm refines the initial detections, ultimately resulting in more precise localization and classification of 3D objects within the scene.
Despite achieving excellent performance, all these methods require costly box annotations, the generation of which is time-consuming and labor-intensive.

\paragraph{3D Object Annotator.} 
Research on reducing annotation costs in 3D object detection tasks has received widespread attention.
To achieve optimal automatic annotation for 3D instances, offboard 3D object detectors \cite{ODNL,offboard,detzero,auto4d, zakharov2020autolabeling} utilize manual annotations and future frame information to train a robust annotator and bounding box corrector. However, the success of offboard detectors still relies heavily on a large amount of manual annotation. 
To reduce dependence, the semi-supervised methods \citep{hssda, detmatch, 3DIoUMatch} select only a small number of fully annotated frames as labeled data, using the remaining frames as unlabeled data. These methods used teacher-student networks for distillation learning to mine and generate pseudo-labels. In addition, sparsely-supervised methods employ an alternative approach to reduce dependency. These methods \citep{CoIn, ss3d,HINTED} adopted a sparsely annotated strategy, retaining only one complete bounding box label for each selected frame. They utilized specially designed unlabeled object mining modules to discover potential pseudo-labels. 
There are also some multimodal weakly supervised object detection methods~\cite{fgr, liu2022multimodal, feng2019deep} that used 2D bounding boxes instead of 3D bounding boxes to carry out object detection tasks.
Although these strategies have significantly reduced the dependence on 3D boxes, it is still not possible to completely abandon laborious box-level annotations.

\begin{figure*}[t]
  \centering
   \includegraphics[width=0.99\linewidth]{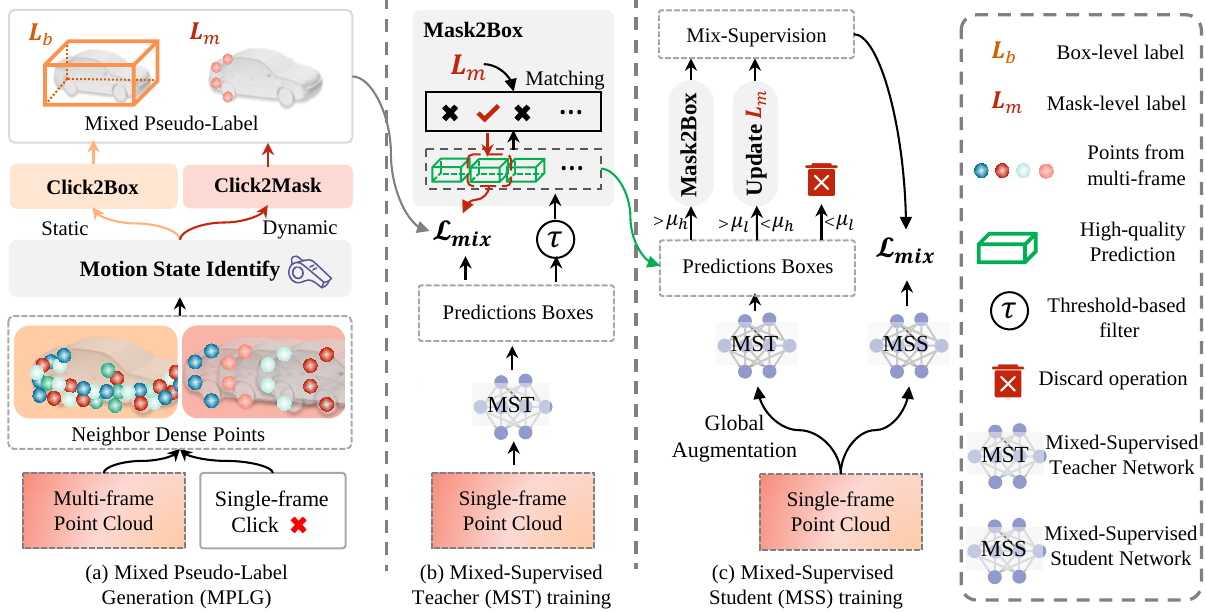}
   \caption{The overview of proposed SC3D. (a) Initially, a novel motion state classification strategy is introduced, followed by the generation of box-level pseudo-label $\textcolor{orange}{\mathbf{L}_{b}}$ and mask-level pseudo-label $\textcolor[RGB]{139,0,0}{\mathbf{L}_{m}}$, utilizing the Click2Mask and Click2Box modules, respectively. (b) With the mixed pseudo-labels generated by stage \textit{(a)}, train the mixed-supervised teacher detector and then update the mask-level supervision to box-level based on high-confidence predictions. (c) Utilizing the generalization of the teacher network to produce mixed pseudo-labels for unlabeled instances, further enhancing the performance of the mixed-supervised student network.}
   \label{fig:framework}
\end{figure*}

\paragraph{Learning From Click-level Annotation.} Click-level annotation is a very labor-saving strategy for manual labeling. Weakly supervised segmentation methods \citep{OTOC, SegGroup, clickseg} have explored the application of only click-level annotation in segmentation tasks. However, this annotation strategy for 3D object detection is still under-explored.
WS3D \citep{WS3D} first introduced click-level annotation into weakly supervised 3D object detection, using partial box-level supervision to guide the inference of click-level annotation towards bounding box estimation. Based on this, ViT-WSS3D \citep{WSS3D} leveraged vision transformer \citep{vit} to further optimize this process. Mixsup \citep{mixsup} made three coarse clicks around an object to generate point cluster labels and proposes a mixed supervision training strategy by combining these cluster-level labels with box-level labels. Current weakly supervised object detection algorithms have successfully incorporated click-level annotations as a substitute for some box-level annotations, but they cannot achieve 3D object detection tasks using only click-level annotations. 
To address this issue, we design a mixed pseudo-label generation module to efficiently convert click annotations into supervision. 

\section{Proposed Approach}

\paragraph{Problem Definition.}
We start by defining the task of weakly-supervised 3D object detection using sparse click supervision. In this setting, the detector is trained exclusively on a single instance-level click annotation $c_{o}=\left \{  x_{o}, y_{o} \right \}$ on each frame of 2D BEV plane, and is then required to predict the full 3D bounding boxes $\left \{ x, y,z,l,w,h,\theta \right \}$. Here, $x, y,z$ represent the coordinates of the 3D center point, while $l, w, h$ denote the object's dimensions, and $\theta$ specifies its orientation. Importantly, the click annotations employed are not limited to precise center clicks; further details are provided in the supplementary material. As shown in Fig.~\ref{fig:framework},  (a) mixed pseduo-label generation; (b) mixed-supervised teacher training; and (c) mixed-supervised student training. The specific steps involved in this process are elaborated as follows.

\subsection{Mixed Pseudo-Label Generation}
To accurately estimate 3D bounding boxes from click annotations as pseudo labels, we aim to leverage temporal cues to enrich sparse point annotations by aggregating registered points across consecutive frames. 
For static instances, the aggregation of dense points facilitates the complete capture of spatial geometric information. However, for dynamic objects, the dense point aggregation often fails to capture the full geometric structure, leading to low-quality pseudo-labels.
This challenge motivates us to first classify the motion status of instances corresponding to each click.

\begin{figure}[t]
  \centering
   \includegraphics[width=0.99\linewidth]{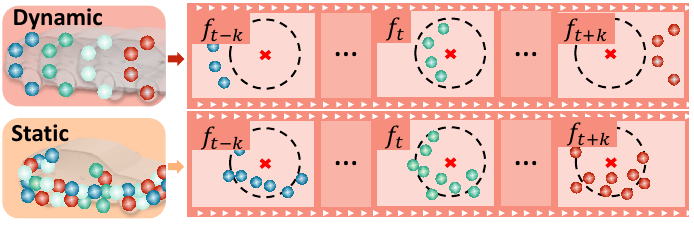}
   \caption{The difference in point distribution between dynamic and static objects in consecutive frames are as follows: dynamic objects have a rapid change in local point density over time; static objects have a stable local point density.}
   \label{fig:sub_ms}
\end{figure}

\paragraph{Motion State Classification for Clicked-instance.} 
As shown in Fig.~\ref{fig:sub_ms}, we observe the duration of local point distribution at clicked positions during a long sequence traversal. 
Specifically, for static instances, the local points at the clicked position exhibits a continuous distribution, whereas, for dynamic instances, the local points is transient throughout the traversal.
Motivated by this, we utilize the persistence of points at local positions within the long sequence for dynamic and static classification.

For each click annotation $c_{o}=(x_{o}, y_{o})$ at the $t$-th frame, we gather adjacent frames $\mathcal{F} = \left \{ f_{t-k},...,f_{t},..., f_{t+k}\right \}$ within a local time window $k$, followed by ground removal~\citep{remove_ground}. Our primary focus is on the BEV points in $\mathcal{F}$, denoted as $ \{\mathbf{P}^{\operatorname{BEV}}_{t} \in \mathbb{R}^{N \times 2}\}_{t\in[t-k, t+k]}$, where $N$ indicates the number of BEV points in each frame. To determine the persistence of the clicked position $(x_{o}, y_{o})$, we search for its neighboring BEV points within a radius $r$, resulting in the collection $\{\mathbf{N}_{t}\}_{t\in[t-k, t+k]}$, with each time step having a cardinality of $N_{t}$:
\begin{equation}
\begin{split}
    \mathbf{N}_{t} = \left\{ p_{i} \in \mathbf{P}^{\operatorname{BEV}}_{t} \mid \left\| p_{i} - c_{o} \right\|_{2} \le r \right\},
N_{t} &= \left| \mathbf{N}_{t} \right|.\\
\end{split}
\end{equation}
To better tally the duration for which points are continuously present near the clicked location, we construct the function $g(t)$, and perform a differentiation operation on $g(t)$:
\begin{equation}
    g(t) = \begin{cases}
 0 & \text{ if } N_{t}= 0; \\
 1 & \text{ otherwise. } 
\end{cases}
\end{equation}

\begin{equation}
    \Delta g(t) = g(t+1) - g(t).
\end{equation}
In adjacent $T$ frames ($T= 2k + 1 $), the duration of the neighborhood points of the click position can be calculated based on the index of the specific value of $\Delta g(t) $. That is, $\Delta g(t)=1 $ indicates the point begins to appear, $\Delta g(t) = -1$ indicates the point disappears, and marking the time difference between the last appearance and the next disappearance of the point on the click annotation frame is the duration time $\Delta t$.
Subsequently, the motion state of the clicked-instance is determined based on the proportion of $\Delta t$ that occupies adjacent $T$ frames.
\begin{equation}
    \begin{cases}
\texttt{static}  & \text{ if } \frac{\Delta t}{T} > \tau;  \\
\texttt{dynamic}  & \text{ otherwise. }
\end{cases}
\end{equation}
where $\tau$ is the duration threshold. If $\frac{\Delta t}{T}$ exceeds the threshold, it indicates that the local point cloud around the click has a longer duration and is considered a static instance. Conversely, if it does not exceed the threshold, it is considered a dynamic instance.
\paragraph{Click2Box.}
For static objects, motivated by~\cite{cpd}, dense points express complete geometric structures, which supports the fitting of high-quality bounding box pseudo-labels from the point cloud distribution. Therefore, we concatenate the neighboring points of multiple frames $\{\mathbf{N}_{t}\}_{t\in[t-k, t+k]}$ to obtain local dense points $\mathbf{D}_t$ for the time step $t$. We perform DBSCAN~\citep{dbscan} clustering algorithm on $\mathbf{D}_t$ to generate several discrete point clusters. 
We retain the cluster of points whose center is closest to the clicked position and consider the points in this cluster as the foreground points of the clicked instance.
Finally, we perform a bounding box fitting algorithm~\citep{L-fit} on the foreground points to generate a box-level pseudo-label.
We utilize Click2Box to infer box-level pseudo-labels \textcolor{orange}{$\mathbf{L}_{b}$} from the click annotations of all static instances.

\paragraph{Click2Mask.} For dynamic objects, we opt to leverage only the single-frame point cloud $\mathbf{P}_{t}$ due to the long-tail distribution observed in aggregated points resulting from motion differences~\citep{mppnet}. Although the instance shape and scale cannot be revealed by click-level labels, the foreground points from click-annotated frame can still provide reliable \textit{semantic} information and coarse \textit{location} information. Consequently, instead of generating box-level pseudo labels for dynamic objects, we produce mask-level pseudolabels by extracting the foreground points in $\mathbf{P}_{t}$. To identify these foreground points, we employ the DBSCAN clustering algorithm on raw point clouds, and select the cluster with the center closest to the click location a mask-level pseudo label. The resulting mask-level pseudo labels denoted as \textcolor[RGB]{139,0,0}{$\mathbf{L}_m$}, are derived from the click annotations of all dynamic instances. 

\subsection{Mixed-Supervised
Teacher Training}

In contrast to traditional 3D object detectors that are solely reliant on box supervision, our approach delves into weakly supervised detectors with mixed supervision. Inspired by MixSup\cite{mixsup}, we re-engineered the strategy for supervision allocation. 
However, the information about mask-level supervision is limited, and it is challenging to achieve optimal detector performance. 
To address this issue, combining with an iterative training, we introduce a mask2box enhanced teacher training strategy, which leverages the high-confidence outputs from the last iterative detector to refine mask-level pseudo-labels into more accurate bounding box-level pseudo-labels, thereby improving the overall precision of the teacher detector.

\paragraph{Mixed Supervised Training.}
To optimize the initial teacher detector, we redesigned the supervision assignment strategy. For all box-level pseudo-labels \textcolor{orange}{$\mathbf{L}_{b}$}, we train the detection network to focus on its position, shape, orientation, and semantics. For mask-level pseudo-labels \textcolor[RGB]{139,0,0}{$\mathbf{L}_{m}$}, the detector focuses solely on semantics and the center of masks. Therefore, the loss function for teacher network can be formulated as: 
\begin{equation}
     \mathfrak{L}_{mix} = \frac{1}{| \mathbf{L}_b |} \sum_{ \mathbf{L}_{b} }\mathfrak{L}_{\operatorname{reg}}+\frac{1}{| \mathbf{L}_{b+m}|} \sum_{  \mathbf{L}_{b+m}   }\mathfrak{L}_{\operatorname{cls}} +  \frac{\lambda}{| \mathbf{L}_m |}\sum_{ \mathbf{L}_m  } \mathfrak{L}_{\operatorname{pos}}.
     \label{eq2}
\end{equation}
$\mathfrak{L}_{\operatorname{reg}}$ and $\mathfrak{L}_{\operatorname{cls}}$ are commonly used regression and classification losses in 3D object detection. $\mathfrak{L}_{\operatorname{pos}}$ is the part that decouples the central position from $\mathfrak{L}_{\operatorname{reg}}$. $\mathbf{L}_{b+m}$ is the union of $\textcolor{orange}{\mathbf{L}_{b}}$ and $\textcolor[RGB]{139,0,0}{\mathbf{L}_{m}}$. Since mask centers are not accurate instance centers, we set the hyper-parameter $\lambda$ to reduce the weight of this part.

\paragraph{Mask2Box Enhanced Teacher.}
Mask-level pseudo-labels provide only semantic and coarse localization information, lacking a description of the instance shape. To compensate for the information lost in mask-level pseudo-labels, we use high-confidence bounding-box predictions of initial teacher to upgrade mask-level pseudo-labels.
Specifically, we first set a threshold $\tau$ to filter out low-confidence predictions from the detector's output. Then, we establish a matching relationship based on the intersection-over-union (IoU) between the $\textcolor[RGB]{139,0,0}{\mathbf{L}_{m}}$ and high-quality prediction boxes $Y$. The bounding box with the highest IoU would replace the $\textcolor[RGB]{139,0,0}{\mathbf{L}_{m}}$ to participate in subsequent training. Mask2Box progressively refines mixed supervision, further enhancing the performance of the teacher network.


\subsection{Mixed-Supervised Student Training}
\begin{figure}[t]
  \centering
   \includegraphics[width=0.99\linewidth]{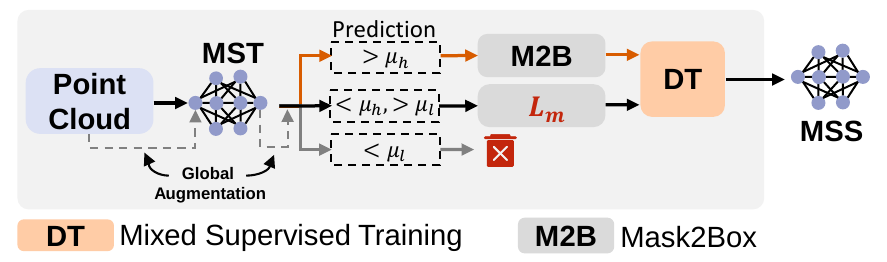}
   \caption{Training strategies for the mixed-supervised student network.}
   \label{fig:teacher-student}
\end{figure}
By progressively recovering from click annotations to supervision signals, we enhance the performance of the teacher model. However, the abundance of unclicked instances within the scenes impedes the training of a well-performed detector. Inspired by ~\cite{HINTED}, we utilize the inherent generalization ability of the teacher network to mine unlabeled instances. Following this idea, ensuring the accuracy of the pseudo-labels output by the teacher model is our primary concern.

\paragraph{Transformation Equivariant Guided Evaluation.} The pseudo-label evaluation solely based on prediction confidence cannot guarantee the reliability of predicted labels~\cite{fixmatch}, which also limits the performance of the teacher network. Inspired by~\cite{TED, ss3d, 3DIoUMatch}, we utilize the transformation equivariant of the detection network to evaluate predicted labels. Specifically, as shown in Fig.~\ref{fig:teacher-student}, we first perform global augmentation (including random rotation, flipping, and scaling) on the input point cloud, and the teacher model produces corresponding detection results $\hat{Y}$. Then, we inversely transform the augmented results and align them back to the initial coordinate system. Based on the property of transformation equivariant, we consider the bounding boxes with high alignment between the $\hat{Y}$ and the original high-quality predicts ${Y}$ as promising pseudo-labels. We use the pairwise IoU scores between $\hat{Y}$ and ${Y}$ as the pseudo-label alignment scores. 

\paragraph{Mixed-supervision Expanding.}
To further refine label quality, we adopt a dual-threshold filter strategy~\cite{hssda}, which uses cluster operation to generate two thresholds $\mu_h$ and $\mu_l$ based on the distribution of IoU scores. For pseudo-labels with alignment scores greater than the high threshold $\mu_h$, we directly utilize them to provide supervision for the training of the student network and serve as support for the Mask2Box module. For pseudo-labels with alignment scores between the two thresholds, they can participate in the training of the student network as mask-level supervision. For pseudo-labels with alignment scores below the low threshold $\mu_l$, we will discard them directly. The training loss of the student network is consistent with that of the teacher network.


\section{Experiments}

\begin{table*}[]
	\centering
\resizebox{1\textwidth}{!}{
	\begin{tabular}{l c c c c c c c c c c c c c}
\toprule
    Method                             & Annotation (cost)      & mAP  & NDS   & Car   & Truck & C.V. & Bus  & Trailer & Barrier & Motor. & Bike & Ped.  & T.C. \\ \midrule 
CenterPoint~\cite{CenterPoint} & Fully Supervised (100\%) & 56.18 & 64.69 & 84.10 & 54.56  & 16.38    & 67.31 & 36.95       & 65.27    & 53.58   & 35.76 & 82.70 & 65.08 \\ \midrule

WS3D~\cite{WS3D} & Boxes(10\%) + Center Clicks  & 39.52 & 47.81 & 78.59 & 42.79  & 5.15    & 38.60  & 17.96       & 46.38    & 27.45   & 16.98  & 75.89 & 45.37
\\
MixSup~\cite{mixsup} & Boxes(10\%) + SAM~\cite{sam}  & 49.49 & 58.65 & 64.63 & 41.71  & 15.61    & 57.57 & 28.19       & 43.56    & 62.28   & 51.42 & 75.07 & 54.87
\\ \midrule

CenterPoint~\cite{CenterPoint} & \multirow{3}{*}{Sparse Boxes (2\%)} & 8.09 & 25.77 & 24.62 & 2.84  & 0.00    & 15.66 & 0.00       & 4.07    & 3.33   & 0.29 & 25.11 & 4.96 \\  
                                  CoIn~\cite{CoIn}&         &    12.47  &   33.79    &  38.70     &    6.85   &   0.00   & 20.67      &   7.81      &    11.51     &   2.85     & 3.36     &   34.85    &  8.50
                                  \\  
                                  HINTED~\cite{HINTED}   &         &    32.62  &  45.76 & 66.63    &  32.71     &    7.59   &   54.56   & 11.42      &   21.16      &    29.12     &   19.08     & 57.22     &   26.63    
                                  \\ \midrule

                            SC3D-Teacher    &   \multirow{2}{*}{Sparse Coarse Clicks (0.2\%)}   & 24.88  & 38.12   & 56.37   &23.47 &1.10 &33.30   & 3.43 & 16.18 &24.67 & 8.81 & 54.42  & 27.00 \\
                             SC3D-Student    &      & 30.71  & 42.37
   & 64.46   & 30.73 & 5.21 & 45.21 & 10.37  & 19.26 & 27.33 & 8.35 & 59.64  & 36.59 \\

         \bottomrule
\end{tabular}
}
\caption{The multi-class results on the nuScenes val set. ‘C.V.’, ‘Ped.’, and ‘T.C.’ are short for construction vehicle, pedestrian, and traffic cones, respectively. 'Boxes + Center-Clicks' denotes that in some scenarios, bounding box annotations are retained, while in the rest, only the central position annotations are preserved. 'Sparse Boxes' indicates that only one bounding box annotation is retained per ten scenes.}
	
    \label{tab:exp2}
\end{table*}

\paragraph{Datasets and Metrics.}
NuScenes~\citep{nuScenes} is a comprehensive autonomous driving dataset that includes 1000 sequences, consisting of 700, 150, and 150 sequences for training, validation, and testing, respectively.  
There are a total of 28k annotated frames for training, 6k for validation, and another 6k for testing. We perform sparse click annotation on the $train$ split of nuScenes to get a new $train$ split.
For evaluation, we adopt mAP and nuScenes detection score (NDS) as the main metrics.

For KITTI dataset \citep{KITTI}, to acquire a dense point cloud, we extracted point cloud data from consecutive frames of the raw data. It is worth noting that the raw data is only used to assist in the generation of pseudo-labels and does not participate in the training process of the detector. For the training stage, following traditional fully-supervised methods~\cite{PV-RCNN, Voxelnet, HANet}, KITTI 3D detection dataset is divided into a $train$ split of $3712$ frames and a $val$ split of $3769$ frames. We perform sparse click annotation on the $train$ split of KITTI to get a new $train$ split. 
To ensure a fair comparison, following sparsely-supervised methods~\citep{ss3d, CoIn}, we calculate the mean Average Precision (mAP) using 40 recall positions with IoU@0.5.  

\paragraph{Implementation Details.}
For coarse click annotations, we use the BEV center of ground truth as a reference and apply a large perturbation range ($0.5 \times w, 0.5 \times l$) to simulate coarse manual clicks. In the supplementary material, we demonstrate a comparison between the effects of simulated clicks and real coarse manual clicks. Meanwhile, in Tab.\ref{tab:exp5}, we analyze the impact of different perturbation ranges on the results. Drawing on human prior knowledge, we establish multi-scale search radii to accommodate objects of different categories. At the training stage, we trained SC3D with a batch size of 32 and a learning rate of 0.003 for eighty epochs on 4 RTX 3090 GPUs. 
More details will be provided in the appendix.

\paragraph{Baselines.}
We are the first to develop a method for training object detectors only with click-annotations, and there are no previously published baselines for comparison. To validate the effectiveness of our SC3D, we chose to compare it with works that are also label-efficient.
To ensure a fair comparison, we adopt widely used VoxelRCNN~\citep{Voxel-RCNN} as basic detector architectural. Furthermore, leveraging the proposed Click2Box module, we infer bounding box supervision from click annotations to construct multiple baselines with click annotations. 
 
 \subsection{Main Results}
 
 \paragraph{Comparison with State-of-the-art Methods.} 
For coarse click annotations, the labeling time per instance is approximately $1.2$ seconds, which is about $2$ times faster than center-click labeling and 100 times faster than bounding box labeling. We primarily conduct comparative experiments with state-of-the-art methods on the nuScenes and KITTI datasets.

At first, we conducted experiments to compare our approach with state-of-the-art label-efficient methods on the nuScenes dataset. To ensure a fair comparison, we follow the previous methods~\cite{CoIn, mixsup} to select the CenterPoint~\cite{CenterPoint} as the base detector. SC3D-Teacher and SC3D-Student, respectively, represent our mixed-supervised teacher detector and mixed-supervised student detector.
As shown in Tab.~\ref{tab:exp2}, SC3D-student achieves performance comparable to the previous best sparsely-supervised method. Meanwhile, the annotation cost of our SC3D-student reduces by more than $10$ times compared to the previously best-performing method. Compared to the precision of the teacher detector, the mAP of the student detector has increased by $5.83\%$, indicating that we have successfully leveraged the teacher detector to mine a large amount of unlabeled instance information. Due to our bounding box pseudo-labels being fitted based on clustering results, our approach performs poorly on larger-scale categories in the nuScenes dataset.


\begin{table}[]
\centering
\resizebox{\linewidth}{!}{
\begin{tabular}{lccllcc}
\toprule
\multicolumn{1}{c}{\multirow{2}{*}{Method}}    & \multicolumn{1}{c}{\multirow{2}{*}{Annotations (cost)}}                                                                                      &  & \multicolumn{3}{c}{3D-Detection}  \\  
\multicolumn{1}{c}{}                        &  &                                                                                   & Easy        & Mod        & Hard             \\ \midrule
Voxel-RCNN~\cite{Voxel-RCNN}           & Fully Supervised                                                                                    100\%                &  & 98.7       & 94.9      & 94.5           \\ \midrule
WS3D~\cite{WS3D}                       & \multirow{2}{*}{Boxes + Center Clicks (3\%)}  &  & 96.3        & 89.4       & 88.9              \\
MixSup~\cite{mixsup}                        &                                                                                               &                        & 94.9        & 92.7       & 90.0             \\ \midrule
SS3D~\cite{ss3d}                             & \multirow{4}{*}{Sparse Boxes (2\%)}                                                                  &  & 98.3        & 89.2       & 88.3             \\
CoIn~\cite{CoIn}                              &                                                                                                                     &  & 96.3        & 86.7       & 74.4             \\
CoIn++~\cite{CoIn}                  &                                                                                               &                        & 99.3        & 92.7       & 88.8              \\HINTED~\cite{HINTED}                 &                                                                                               &                        & 98.5        & 91.6       & 90.3                   \\ \midrule
 SC3D-Teacher                                   & \multirow{2}{*}{Sparse Coarse Clicks (0.2\%)}                                                                                             &  &94.7        & 87.4       & 80.3             \\ 
 SC3D-Student                                  &                                                                &  &96.4        & 91.6       & 84.6        
\\
\bottomrule
\end{tabular}}
\caption{Experimental results on KITTI dataset compared with recent state-of-the-art label-efficient methods. We report results of car with 40 recall positions, below the 0.5 IoU thresholds. 
}

\label{tab:exp1}
\end{table}

For KITTI dataset, following the mainstream approaches~\cite{ss3d, CoIn, HINTED}, we also adopted Voxel-RCNN~\cite{Voxel-RCNN} as the base detector. Since our initial bounding box labels are generated based on point cloud distribution and significantly differ from manual annotations, we refer to~\cite{MODEST} and compare the evaluation results under the $0.5$ IOU threshold.
Despite employing a more lightweight annotation form, retaining only coarse click annotation, our SC3D still achieves comparable performance with other methods. Furthermore, with only $0.2\%$ of the labeling cost, the proposed SC3D achieves $94\%$ of the average performance of full supervision.

\paragraph{Comparison with the mixed-annotation method.} 
\begin{table}[]
\centering
\resizebox{\linewidth}{!}{
\begin{tabular}{cccccccc}
\toprule
Method & Annotation                     & \multicolumn{3}{c}{3D AP@0.7} & \multicolumn{3}{c}{BEV AP@0.7} \\ \midrule
WS3D   & clicks +  $ precisely^\# $      & 84.0    & 75.1    & 73.2   & 88.5    & 84.9    & 84.7    \\
SC3D   & s-clicks  & 69.3    & 56.0    & 50.0   & 89.1    & 76.9    & 69.6    \\
SC3D   & s-clicks + $precisely^\#$ & 91.6    & 77.4    & 74.3   & 95.4    & 86.1    & 83.3    \\\bottomrule
\end{tabular}
}
\caption{Comparison with WS3D on KITTI $val$ split. We report results of car with 40 recall positions, below the 0.7 IoU thresholds. \enquote{s-clicks} denotes sparse coarse clicks. \enquote{$ precisely^\# $} denotes 534 precisely-annotated instances.}
\label{tab:mix}
\end{table}

To reduce the impact of manual annotation habits on label quality, we refer to mixed annotation strategies\cite{WS3D}, incorporating a small number of additional manually annotated bounding boxes during detector training. As shown in Tab.~\ref{tab:mix}, with the same number of precisely annotated instances, our approach can achieve better performance than WS3D at an IoU threshold of 0.7 using fewer clicks. This proves that SC3D can learn human annotation habits with a small amount of manual annotation, ultimately reaching the goal of adjusting bounding boxes.

\paragraph{Comparison with Full Clicks.} To validate the detection performance of SC3D on unlabeled instances, we introduced a new click setting for performance comparison: Full Coarse Clicks, which involves annotating clicks for all instances in the nuScenes and KITTI $train$ splits.
Tab.~\ref{tab:sparse_full} displays the experimental results conducted with two distinct click annotation schemes. From the experimental results, it is evident that compared to full click annotations, SC3D achieves similar performance under sparse annotations, especially on the KITTI dataset, with a precision loss of no more than $2\%$.

\begin{table}[]
\centering
\resizebox{\linewidth}{!}{
\begin{tabular}{lcllll}
\toprule
\multirow{2}{*}{Method} & \multirow{2}{*}{Annotation} & \multicolumn{2}{c}{nuScenes} & \multicolumn{2}{c}{KITTI} \\
                        &                             & mAP           & NDS          & 3D    & BEV       \\ \midrule
\multirow{2}{*}{SC3D}   & Full Coarse Clicks          &      44.0         &       49.8       &  92.5       &  94.1           \\
                        & Sparse Coarse Clicks        &   30.7            &   42.3           &    91.6      &   92.0        \\ \bottomrule
\end{tabular}
}
\caption{Comparison with the results of different click cost. Full Coarse Clicks refers to the rough annotation of all instances in the scene through coarse clicking. We report the results of mAP and NDS for the nuScenes dataset, as well as 3D@0.5 and BEV@0.5 for the KITTI dataset.
}
\label{tab:sparse_full}
\end{table} 

\begin{figure*}[h]
  \centering
  \begin{subfigure}[b]{0.32\textwidth}
    \includegraphics[width=\textwidth]{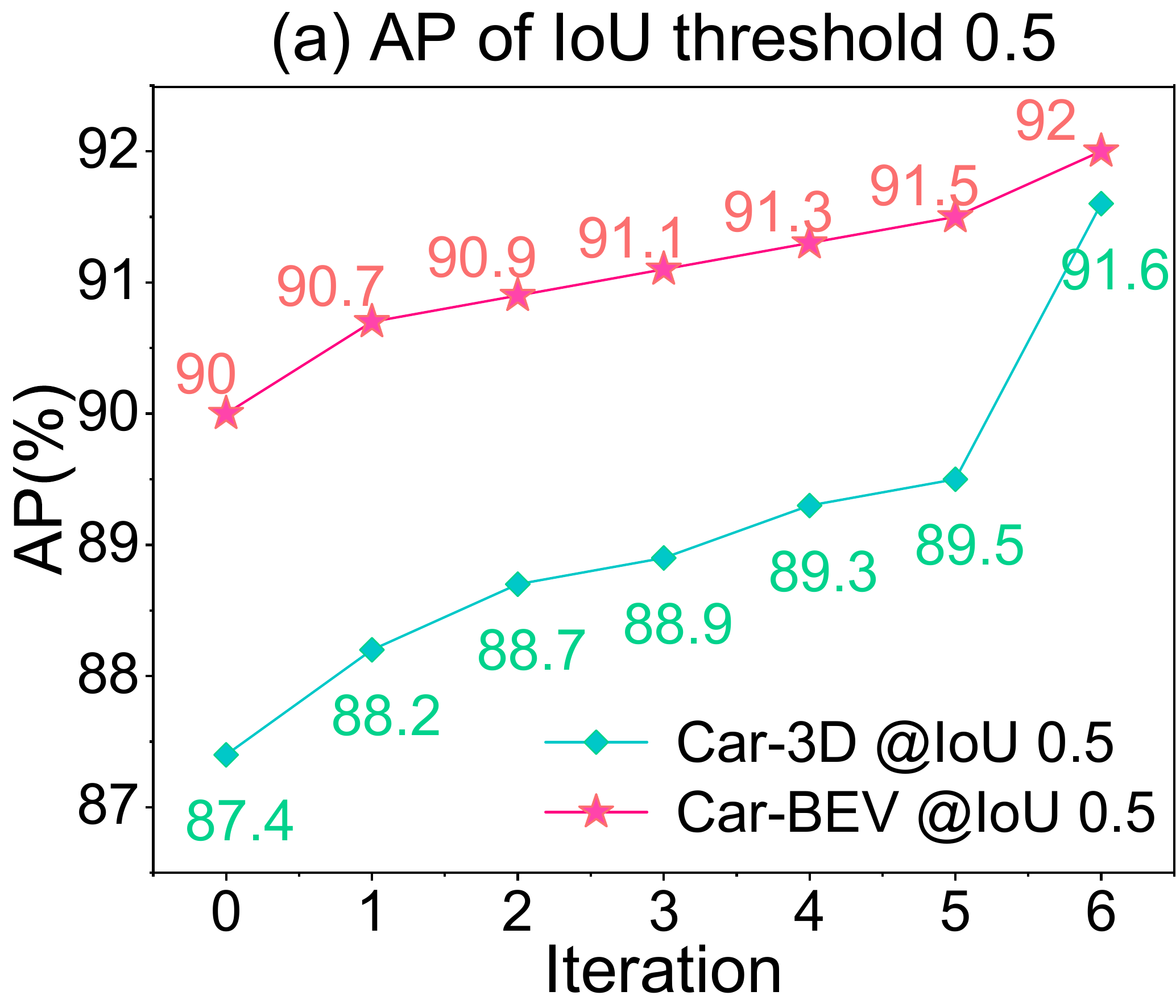}
    \label{fig:sub1}
  \end{subfigure}
  \hfill 
  \begin{subfigure}[b]{0.32\textwidth}
    \includegraphics[width=\textwidth]{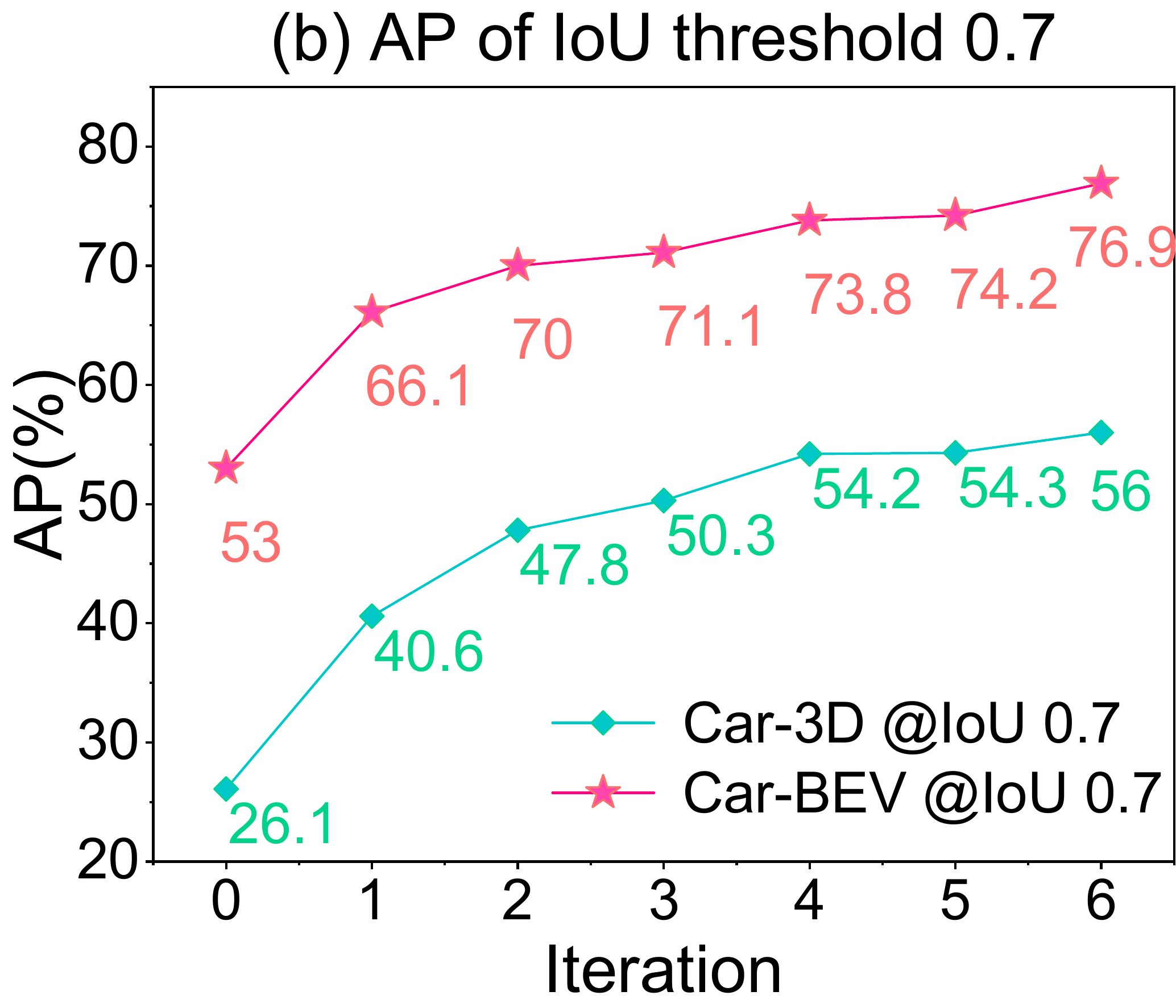}
    \label{fig:sub2}
  \end{subfigure}
  \hfill 
  \begin{subfigure}[b]{0.32\textwidth}
    \includegraphics[width=\textwidth]{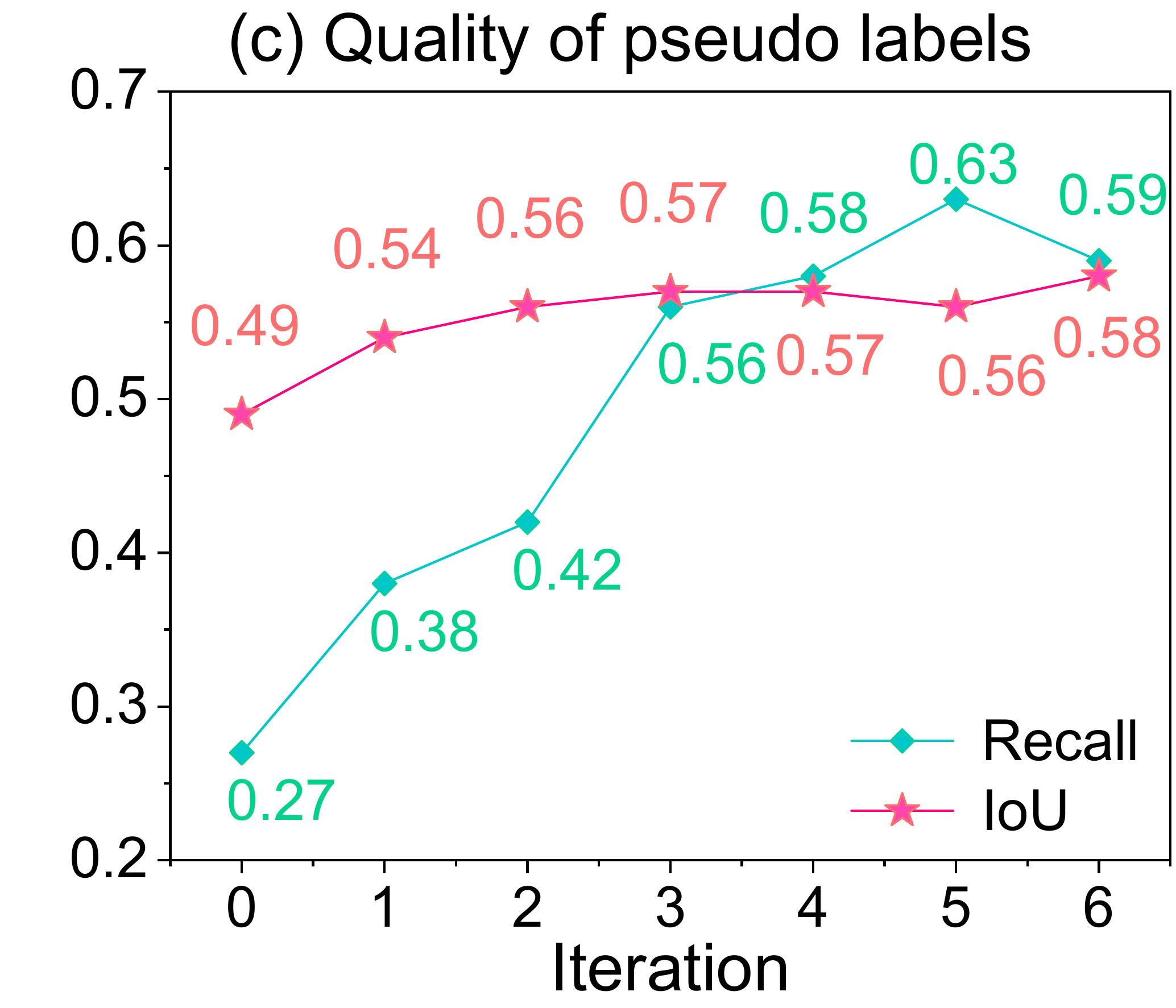}
    \label{fig:sub3}
  \end{subfigure}
  \vspace{-4ex}
  \caption{Prediction results and the quality of pseudo-labels across various iterative rounds. (a) and (b) represent the 3D and BEV results for cars at IOU thresholds of 0.5 and 0.7, respectively; (c) compare the pseudo-labels with the ground truth across each iterative round. }
  \label{fig:finalfigure}
\end{figure*}

\subsection{Ablation Study and Analysis}
In this section, we conduct ablation studies to evaluate our proposed modules and parameters in SC3D. Models are trained on KITTI's training set and tested on its validation set. We use Voxel-RCNN for our study due to its quick training, and results with other detectors are comparable.

\begin{table}[h]
\centering
\resizebox{\linewidth}{!}{
\begin{tabular}{ccccccc}
\toprule
\multirow{2}{*}{MPLG} & \multirow{2}{*}{Mask2Box} & \multirow{2}{*}{MSS} & \multicolumn{2}{c}{Car AP@0.5} & \multicolumn{2}{c}{Car AP@0.7} \\
                      &                           &                      & 3D             & BEV           & 3D             & BEV           \\ \midrule
                    &                           &                      & 69.5           & 79.3         & 17.6           & 49.8          \\
            \checkmark          &                           &                      & 87.4           & 90.0          & 26.1           & 53.0          \\
            
           \checkmark           &            \checkmark               &                      & 89.4           & 91.6          & 53.0           & 73.9          \\
              \checkmark        &   \checkmark                        &        \checkmark              & 91.6           & 92.0          & 56.0           & 76.9          \\ \bottomrule
\end{tabular}
}
\caption{Ablation study under different module.  We report the car-mod. results with 40 recall positions, under 0.5 and 0.7 IoU thresholds.
}
\label{tab:ablation}
\end{table}

\paragraph{Effect of Mixed Pseudo Label Generation (MPLG).}
Tab.~\ref{tab:ablation} presents the results of testing each module individually. The first row shows the results of generating box-level pseudo-labels directly from click annotations using the unsupervised approach~\cite{OYSTER}. Due to the difficulty in recovering the geometric structure of moving objects across consecutive frames, naive label generation methods struggle to ensure label quality. The second row demonstrates the outcomes of our proposed MPLG module, significantly enhancing the performance of the baseline method. This confirms the effectiveness of MPLG's strategy for generating corresponding labels based on the motion status of objects.

\paragraph{Effect of Mask2Box.}
Although the mask-level labels generated by MPLG can convey supervision information about moving objects, this descriptive form still cannot fully express the structure of moving objects. To address this issue, we designed the Mask2Box module to enhance mask-level pseudo-labels. The results in the third row of Tab.~\ref{tab:ablation} validate the effectiveness of Mask2Box, particularly with a substantial increase in accuracy at an IoU threshold of $0.7$.

\paragraph{Effect of Mixed-Supervised Student (MSS) training.}
The teacher-student network has been proven to be a viable solution for addressing sparsity issues~\cite{HINTED}. Therefore, we propose a mixed-supervised teacher-student network for sparse click strategies, where the teacher detector leverages transformation equivariance to provide high-quality labels for the student network. The last row of Tab.~\ref{tab:ablation} demonstrates the results of the student network, indicating that the teacher-student network architecture has further enhanced the performance of the detector.

\paragraph{Effects of Perturbation Factor $\delta$.}
\begin{table}[h] 
    \centering
    \resizebox{0.9\columnwidth}{!}{
    \begin{tabular}{ccccccc}
\toprule
\multirow{2}{*}{$\delta$} & \multicolumn{3}{c}{Car-3D AP@ 0.5} & \multicolumn{3}{c}{Car-BEV AP@ 0.5} \\ 
                                          & Easy       & Mod.       & Hard      & Easy       & Mod.       & Hard      \\ \midrule
0.25                                      & 96.6       & 92.2       & 87.0      & 96.6       & 92.4       & 87.3      \\
0.5                                       & 96.4       & 91.6       & 84.6      & 96.5       & 92.0       & 84.9      \\
0.7                                       & 96.0       & 89.5       & 82.2      & 96.1       & 91.5       & 84.4      \\
1.0                                       & 95.0       & 87.4       & 80.2      & 95.1       & 87.7       & 80.7      \\ \bottomrule
\end{tabular}
}
\captionof{table}{Effects of perturbation factor $\delta$. We report results on the Car-3D and Car-BEV detection with 40 recall positions, below the 0.5 IoU threshold. }
    \label{tab:exp5}
\end{table}

Tab.~\ref{tab:exp5} demonstrates the robustness of the coarse click.
To rigorously verify the impact of the perturbation factor, we conduct experiments on the more challenging sparse clicking setup. 
We applied various disturbance factors $\delta = 0.25, 0.5, 0.7, 1.0$, defining the disturbance range as $(\delta \times w, \delta \times l)$
, where $w$ and $l$ are the size of the object. Although the performance of the detector experiences a slight decline with the increase of perturbation, the overall performance remains relatively stable.
Even with a large disturbance range ($\delta = 1.0$), our SC3D still achieves satisfactory performance.

\paragraph{Weight Selection for $\lambda$.} 
In Tab.~\ref{tab:exp6}, we delve into the discussion of the hyper-parameter $\lambda$ as presented in Equation \eqref{eq2}, specifically in the context of our model without the incorporation of the Mask2Box modules. The results of Tab.~\ref{tab:exp6} reveal that assigning lower weights to predictions with ambiguous locations leads to improved outcomes. This observation suggests that, in the case of ambiguous location supervision, supplying a single, subtle cue is adequate to attain peak detection performance. Conversely, assigning excessive weight to such ambiguous cues can introduce unwanted noise into the detection process, potentially degrading the model's accuracy and reliability. 


\begin{table}

    \centering
    \resizebox{0.9\columnwidth}{!}{
    \begin{tabular}{ccccccc}
\toprule
\multirow{2}{*}{$\lambda$} & \multicolumn{3}{c}{Car-3D AP@ 0.5} & \multicolumn{3}{c}{Car-3D AP@ 0.7} \\  
                                          & Easy       & Mod.       & Hard      & Easy       & Mod.       & Hard      \\ \midrule
0.2                                      & 94.6
       & 87.7       & 82.8      & 59.8       & 51.0       & 46.0      \\
0.5                                       & 94.5       & 87.4
       & 82.8      & 57.5
       & 48.4       & 45.0
      \\
0.7                                       & 93.8
       & 87.3
       & 82.7
      & 56.2

       & 47.8
       & 44.7
      \\
1.0                                       & 94.6
      & 87.3       & 82.5
      & 55.8
       & 47.7       & 44.0
      \\ \bottomrule
\end{tabular}
}
\captionof{table}{Weights Selection for $\lambda$. We report results with 40 recall positions, below the 0.5 and 0.7 IoU thresholds. }
\label{tab:exp6}
\end{table}

\paragraph{Prediction Results and Quality of Pseudo-labels at Each Iteration.}

In Fig.~\ref{fig:finalfigure} (a) and (b), we show the performance of SC3D over different numbers of updates and iterations in the teacher-student network. 
At both the 0.5 and 0.7 thresholds, the performance is significantly enhanced.
The improvement may be attributed to the student network receiving more information from unlabeled instances during the iterative updates. 
To verify this hypothesis, we use the ground truth as a reference, obseving the average IoU and recall throughout the iterative learning process. 
As shown in Fig.~\ref{fig:finalfigure} (c), there is an overall stable upward trend in the quality of our pseudo-labels as the number of iterations increases. The increase in Recall confirms the successful mining of unlabeled instances. 

\section{Discussion and Conclusion}
We designed an efficient annotation strategy, \textit{single click per frame}, tailored for label-efficient 3D object detection and proposed a weakly-supervised object detection method, SC3D, leveraging this approach. SC3D is mainly divided into three parts: mixed label generation, mixed teacher model training, and mixed student model training.
In the mixed pseudo-label generation stage, we propose a novel method for dynamic and static classification, and design the Click2Box and Click2Mask modules according to the motion states of objects to generate mixed pseudo-labels.
Subsequently, we propose a mixed-supervised teacher model that empowers the detector to assimilate a blend of supervision signals. Ultimately, we introduce a mixed-supervised student network, which leverages the teacher model's generalized insights to learn from unclicked instances.
Extensive experiments on the nuScenes and KITTI datasets have demonstrated that our SC3D method, using only sparse clicks and incurring a labeling cost of just 0.2\%, can achieve commendable performance.

\paragraph{Limitations.} Although the mixed pseudo-label generation stage provides the detector with good initial pseudo-labels that describe the geometric shape and location of instances, these rule-based labels may not match the high-quality annotations produced by human annotators. 
This discrepancy leads to reduced performance under higher IoU thresholds, where precise alignment is critical. 

{
    \small
    \bibliographystyle{ieeenat_fullname}
    \bibliography{main}
}


\end{document}